\documentclass[11pt,a4paper]{article}
\usepackage[hyperref]{acl2020}
\usepackage{times}
\usepackage{latexsym,booktabs, afterpage}
\usepackage{graphicx}

\usepackage{microtype}

\aclfinalcopy 
\def\aclpaperid{4} 

\title{Screenplay Quality Assessment: Can We Predict Who Gets Nominated?}

 \author{Ming-Chang Chiu ~~ Tiantian Feng ~~ Xiang Ren ~~ Shrikanth Narayanan\\
   Department of Computer Science\\
   University of Southern California \\
   Los Angeles, CA 90089, USA\\
   \texttt{\{mingchac,tiantiaf,xiangren\}@usc.edu}\\ \texttt{shri@sipi.usc.edu}}

\begin{document}
\maketitle

\begin{abstract}
Deciding which scripts to turn into movies is a costly and time-consuming process for filmmakers. Thus, building a tool to aid script selection, an initial phase in movie production, can be very beneficial. Toward that goal, in this work, we present a method to evaluate the quality of a screenplay based on linguistic cues. We address this in a two-fold approach: (1) we define the task as predicting nominations of scripts at major film awards with the hypothesis that the peer-recognized scripts should have a greater chance to succeed. (2) based on industry opinions and narratology, we extract and integrate domain-specific features into common classification techniques. We face two challenges (1) scripts are much longer than other document datasets (2) nominated scripts are limited and thus difficult to collect. However, with narratology-inspired modeling and domain features, our approach offers clear improvements over strong baselines. Our work provides a new approach for future work in screenplay analysis. 
\end{abstract}

\section{Introduction}

The motion picture industry is a multi-billion dollar business worldwide \cite{michael_screenplay}. Decisions in selecting movies to be produced are critical to the profitability of a movie studio. However, the selection of the screenplay that happens at the initial phase of a movie production pipeline and has a large influence on the financial budget and quality of the final movie production, has a large subjective element. 
Thus, an objective and reliable tool to help evaluate and narrow down the candidate scripts is of vital importance to aid the ``green-lighting" (deciding which scripts to turn into movies) process. 


In general, movie script writing can follow a well-defined \textit{Three Act} structure \cite{field2007, mckee1997}. Also, Weiland \cite{weiland2013structuring, weiland2018structure} specifies a more fine-grained storytelling plan, starting from \textit{hook, inciting event, 1st plot point, 1st pinch point, midpoint, 2nd pinch point, 3rd plot point, climax} to \textit{resolution}, what are called Structural Points (SP). We believe knowledge like the above in strucuring a screenplay can bring benefits in selecting the most relevant textual properties for the prediction of script quality. 

Aside from the event positioning, \citet{follows2019scriptcoverage} reported that how writers develop characters and events, i.e., \textit{Characterization} and \textit{Plot}, are two main foci of industry reviewers. We thus devise our domain specific features in these two aspects. We hope to offer an enhanced understanding of the essential elements in high-quality movie scripts. 

To perform quality assessment, based on an assumption that the nominated scripts are recognized writings and thus should have had higher chance of passing green-lighting, we propose to perform an evaluation in a two-fold approach. First, we use award-nomination prediction as a proxy to the green-lighting process. Second, we examine our domain features and models by integrating them into existing document classification methods.

We acknowlodge the constraints of our metric in that the number of award venues has its limits, and not necessarily those without nomination would be any worse than the nominated. But due to the difficulty in collecting unproduced scripts with peer reviews, we adopt our current approach.


Our main contributions are as follows: (1) We defined a quality metric for screenplays and collected ground truths from peer-reviewed venues. (2) Based on structural knowledge of screenplay narratology, we developed a simple narratology-inspired model for our task. (3) Motivated by industry opinions and narratology, we devised domain-specific features to achieve our objective. (4) We tested that for long document classification, a simple feature-based approach can work better than state-of-the-art models.

\section{Related Works}
\label{sec:related}

Literary works-related research has gained interest in recent years. \citet{bamman-etal-2013-learning, bamman-etal-2014-bayesian} have succeeded to learn latent character types in film and novels;  \citet{iyyer-etal-2016-feuding, chaturvedi2016modeling, elson-etal-2010-extracting} try to model character relations in novels. \citet{Papalampidi2019MoviePA} analyze narrative structure of movies by using turning points, and \citet{chambers-jurafsky-2008-unsupervised, sims-etal-2019-literary} seek to detect events in narratives.


A noteworthy attempt in measuring quality of literary works we know of is made by \citet{kao-jurafsky-2012-computational}, who quantitatively analyze various indicators for discerning professional poems from amateurs'. However, in script writing, the cinematic success criteria lack evaluative consensus \citep{Simonton2009CinematicSA} --- previous works on evaluation of movies have largely focused on forecasting revenue or profit of movies using production, distribution, and advertising data \citep{GHIASSI20153176, Lash2015EarlyPO} or basic textual and human annotated features \citep{eliashberg_box_prediction}. 

The main differences between our work and previous works are: (1) our approach aims to process automatically without human annotated features. (2) our metrics and methods are geared towards evaluation that based solely on textual properties.

\section{Data and Problem Setting}

\noindent
\textbf{Data collection.} We evaluated our method using ScriptBase \cite{Gorinski2018WhatsTM} and Movie Screenplay Corpus (MSC) \citet{Ramakrishna2017LinguisticAO} datasets. ScriptBase provides 917 scripts and MSC contains 945 Hollywood movies. We kept 897 and 868 suitable ones which have enough character utterances for our approach from each dataset respectively. Similar to \citet{underwood2019distant}, which analyzes high-prestige novels as works that have been reviewed by top journals, we collected the screenplays that have histories of nominations as quality ``ground truth". The venues we collect from are well-known professional prizes, which include ``Writers Guild of America Award", ``Academy Awards", ``Golden Globe Awards", and ``British Academy of Film and Television Arts Awards". Since we focus on textual properties for success, we only gleaned nominations in the original screenplay and adapted screenplay categories. In the end, we obtained 212 (23.6\%) movies out of ScriptBase and 113 (13.0\%) from MSC as quality ``ground truth" labels.

\noindent
\textbf{Problem Setup.} Our work focuses on measuring quality as whether or not a movie would be nominated at a peer-reviewed venue. The basic assumption for using this approach as success metrics is simple --- a screenplay that receives nominations by critical reviewers should have had higher chance of getting through green-lighting. 

\noindent
\textbf{Challenges.} By nature, a movie should be tough to be cleanly categorized, due to its length, complex storyline and turns, and the lack of evaluative criteria. Prior works in document classification \cite{Yang2016HierarchicalAN, Liu2017DeepLF, Adhikari2019RethinkingCN, johnson-zhang-2015-effective} evaluated on datasets with small document size (Reuters, IMDB, Yelp, etc.). However, our document size on average is at least 65 times longer, which may be challenging for NN-based models to train due to long sequences and the associated computational burden. Besides, the number of training data we have is at most 1000 times smaller than other datasets. With our datasets being \textbf{long}, \textbf{fewer} and \textbf{skewed}, state-of-the-art deep learning techniques may not work well. Summary of the comparisons is shown in Table~\ref{tab:dataComp}. 

\begin{table}
\centering
\scalebox{0.9}{
\begin{tabular}{lrrc}
    \toprule
    Dataset  & documents & average \#w  & \%pos \\ \midrule
    Reuters & 10,789 &  144.3 & -\\ 
    IMDB & 135,669 & 393.8 & -\\ 
    Yelp 2014 & 1,125,457 &  148.8& - \\ \midrule
    ScriptBase & 897 & 27,539.7 & 23.6\\
    MSC & 868 & 27,067.4 & 13.0\\\bottomrule

\end{tabular}
}
\captionof{table}{\textbf{Dataset statistics and comparisons of datasets.} \#w denotes the number of
words and \%pos denotes the percentage of positive class.}
 \label{tab:dataComp} 
\end{table}

\section{Analysis of Domain Features}
\label{sec:features}

In this section, we introduce our domain features that are divised to achieve our goal and provide analysis based on our problem setup.

\textit{Characterization} and \textit{Plot} are major aspects of focus in the industry; inspired by which, we devised 6 novel features. For each, we provide intuitive motivations, and then detail how we converted them computationally. We chose the top two most speaking characters of each movie to analyze for \textit{characterization}.

According to \citet{weiland2018structure}, a script can place 9 SPs roughly equally distributed, creating eight equal-lengthed development segments (DS) in between. We hypothesize that such structural hints should help to achieve our objective. Based on the statistics of both datasets, to leverage the SPs, we collected a context window of 1\% ($\sim$270 words) centered at SPs for all scripts. 


By the definition of characterization, we hypothesized that by measuring pattern change of characters, we may see how writers develop the characters' personality. We sought pattern change via two kinds of changes writers would make between SPs - linguistic (speaking pattern) change and emotional change. To do this, we proposed \textit{Linguistic \& Emotional Activity Curve}. 

\noindent
\textbf{Linguistic \& Emotional Activity Curve (\textit{ling, emo}).} For linguistic change, we extracted the dependency trees of characters; for emotional change we used normalized Empath \citep{Fast2016EmpathUT} to get characters' emotion status. We combined the linguistic distribution, Empath distribution of sentences in each DS with \textit{activity curve} \cite{dawadi2016modeling}, which uses a Permutation-based Change Detection in Activity Routine (PCAR) algorithm, to measure the change between two DSs of distributions.

\noindent
\textbf{Type-token ratio (\textit{tt}).} As \citet{kao-jurafsky-2012-computational} show, in poetry, the \textit{type-token ratio} related most positively to the quality of a poem. We believed this concept should work similarly on character analysis, and can show how much effort writers devoted in characterization. We defined this feature as the number of unique words used by a character divided by the total number of words.

\noindent
\textbf{Valence-Arousal-Dominance (\textit{VAD}).} \citet{vad-acl2018} performed extensive study in getting an objective score for words in VAD dimensional space \citep{Russell1980ACM, Russell2003CoreAA}. We used average scores over the context window of each SP to represent level of emotion. 

\noindent
\textbf{Emotion Intensity (\textit{int}).} Similar to \textit{VAD}, we used the NRC Affect Intensity Lexicon \citep{LREC18-AIL} over the SPs to score emotion intensity along four basic emotion classes \citep{PLUTCHIK19803}.

Also, since events are usually addressed in units of scenes, we wanted to get a picture of how many different emotionally similar scenes across the dataset appear in a movie.

\noindent
\textbf{Empath Clustering (\textit{clus}).} We retrieved lexical categories for each uttrance from Empath and then clustered the lexical category distributions of all utterances with deep embedded clustering \cite{xie2016unsupervised}. We obtained the cluster distribution based on the lexical categories within a movie as a feature representation.

\begin{figure}
\begin{minipage}[a]{1\linewidth}
  \centering
  \centerline{\includegraphics[width=8.5cm]{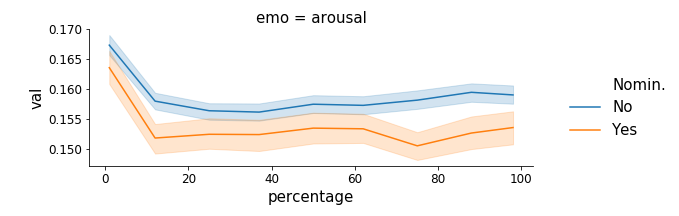}}
\end{minipage}

\begin{minipage}[b]{1\linewidth}
  \centering
  \centerline{\includegraphics[width=8.5cm]{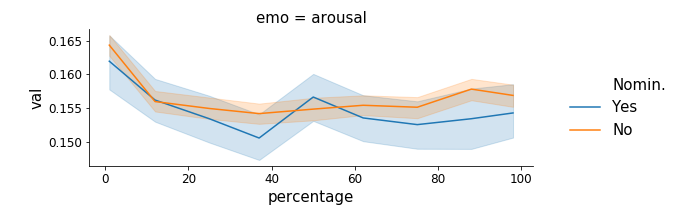}}
\end{minipage}
\caption{\textbf{Nomination vs Non-nomination of arousal level along percentage of scripts.} \textit{Upper:} ScriptBase. \textit{Lower:} MSC.}
\label{fig:arousal}
\end{figure}

We visualized partial features in a ``nomination vs non-nomination" fashion, as in Fig.~\ref{fig:arousal}, to show the potential of our features. For some we can easily observe clear differences from one to the other, while some are more subtle. For instance, in \textit{VAD}, the \textit{arousal} of MICA is ambiguous between the two, and yet we can easily discern nominated scripts along the same axis for ScirptBase.

\section{Predictive Modeling}
\label{sec:domainmodel}
In this section, we define our prediction task, and then propose our base model and then move on to a paradigm which integrates domain features proposed in previous section.

\noindent
\textbf{Task Formulation.} 
As a proxy to the original quality assessment task, we define a binary classification task as to predicting the nomination of a script. 

\begin{figure}[ht]
    \centering
    \includegraphics[width=0.5\textwidth]{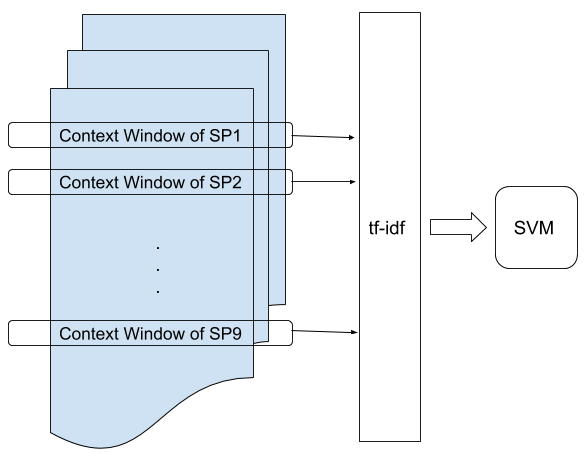}
    \caption{\textbf{Narratology-inspired model workflow.}}
    \label{fig:tfidfSVM}
\end{figure}

\noindent
\textbf{Narratology-inspired Model.} Inspired by narratology, 
we propose \textit{Tfidf-SVM$_{narr}$} --- instead of using all texts in an entire document, we extract words in context window of SPs for each document, compute the tf-idf representations, and feed them into a SVM classifier. The main components of \textit{Tfidf-SVM$_{narr}$} are shown in Fig.~\ref{fig:tfidfSVM}. Due to the large amount of unique tokens, we chose only the top 500 important features to represent a document. We test the results without choosing 500 features and our setting is better.

\noindent
\textbf{Feature-based Prediction.} To examine the predictive power of proposed features, on top of 
\textit{Tfidf-SVM$_{narr}$}, we add domain features along with tf-idf to SVM to see the efficacy of domain features.  


\section{Experimental Setups}


\noindent
\textbf{Dataset usage.} We performed random sampling on both datasets such that 80\% is used for training, 10\% for validation, and 10\% for test.

\noindent
\textbf{Baselines.} We adopted HAN \citep{Yang2016HierarchicalAN}, BERT$_{base}$, BERT$_{large}$ \citep{Devlin2019BERTPO} as our baselines. Since a script is subdivided into scenes, our HAN implementation, HAN$_{scence}$, uses scence as the second hierarchy instead of sentence.

\noindent
\textbf{Implementation details.} We use Scikit-learn 0.21.3 to implement feature-based models, and PyTorch 1.3.1 for deep neural models. With Hugginface \citep{Wolf2019HuggingFacesTS}, we overcome BERT's 510-token limit by applying averaging pooling on the sequence of BERT $h_{[CLS]}$ hidden states of subchunks of the script to get a global context vector, and then fine-tune the task end-to-end. And since the binary labels in both datasets are imbalanced, we weight the positive class by inverse frequency of class labels in the training set. 

\noindent
\textbf{Hyper-parameters.} To ensure a fair comparison, we tuned the hyper-parameters for all models. On feature-based models, we performed grid search. For NN models, we use embedding size 100 and Adam optimizer with 0.001 learning rate.

\section{Results and Discussion}

We report the macro-averaged F1 scores of each model in Table~\ref{tab:summary}, interestingly, from which we see that NN-based document classification methods are no better than our proposed simple narratology-based model. We suppose the length of document could be the main reason, RNNs or transformers may not handle ``super long-term depdendencies" well for complex compositions like movie scripts.

\begin{table}
\centering
\scalebox{0.75}{
\begin{tabular}{lcc}
    \toprule
    \textbf{Method}~/~\textbf{Dataset} & \textbf{ScriptBase} & \textbf{MSC} \\
    \midrule
    {HAN$_{scene}$} & 45.12 & 45.62 \\ 
    {BERT$_{base}$ }& 42.67 & 46.29 \\ 
    {BERT$_{large}$} & 42.67 & 46.29 \\ 
    Tfidf-SVM & 47.01 & 59.21 \\ \midrule
    {TFIDF-SVM$_{narr}$} & \textbf{57.43} & \textbf{59.21} \\
    { + emo + VAD} & 56.52 & 55.29 \\
    { + ling + emo + tt} & \textbf{62.35} & 62.73 \\
    { + int + ling + emo + clus} & 60.87 & \textbf{64.79} \\
    \bottomrule
\end{tabular}
}
\caption{\textbf{F1 scores (\%) of model predictions.} }
\label{tab:summary}
\end{table}

\begin{figure}[ht]
    \centering
    \includegraphics[width=0.5\textwidth]{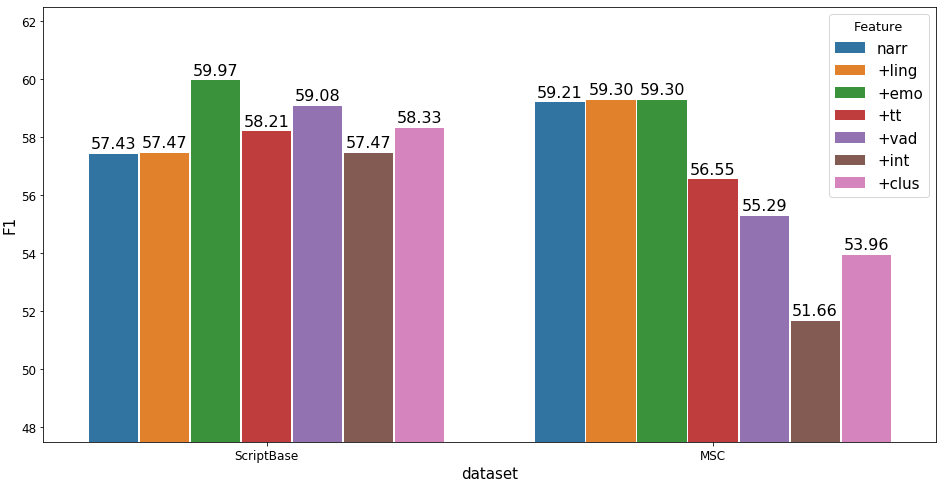}
    \caption{\textbf{Individual feature effect.} F1 scores of \textit{Tfidf-SVM$_{narr}$} and adding proposed features individually.}
    \label{fig:f1_bar}
\end{figure}

In Fig.~\ref{fig:f1_bar}, we show the effect of each individual feature. \textit{Linguistic \& Emotional Activity Curve} show improvements on both datasets, and yet the rest do not consistently help, especially on MSC, we think it may be because (1) the tfidf has 500 dimensions so individual feature may be overwhelmed, but, more features combined such as adding \textit{int+ling+tt} can generate consistent improvements, (2) the efficacy of feature can be dataset-dependent, e.g., we do not observe significant differences in \textit{Arousal} of MSC as in its ScriptBase counterpart (Fig.~\ref{fig:arousal}), and so does the classifier. Besides, adding features with negative correlations can damage the performance, e.g., adding \textit{emo \& vad}.

\section{Conclusion}
\label{sec:conclusion}

We present a novel approach and features to systematically analyze the quality of a screenplay in terms of its festival nomination-worthiness. This can serve as a preliminary tool to help filmmakers in their decision-making, or on the other hand, an objective way for writers to compare their works with others. Our results also show that simple lightweight approach can outperform state-of-the-art document classification methods. This also points out the current deficiency for long document classification research in the community.


\bibliography{anthology,acl2020}
\bibliographystyle{acl_natbib}

\end{document}